\documentclass{article}

\PassOptionsToPackage{numbers}{natbib}

\usepackage[final]{neurips_2022_ml4ad}

\usepackage[utf8]{inputenc} 
\usepackage[T1]{fontenc}    
\usepackage{hyperref}       
\usepackage{url}            
\usepackage{booktabs}       
\usepackage{amsfonts}       
\usepackage{nicefrac}       
\usepackage{microtype}      
\usepackage{xcolor}         
\usepackage{float}
\usepackage{graphicx}
\usepackage{verbatim}       

\title{Improving Motion Forecasting for Autonomous Driving with the Cycle Consistency Loss}

\author{%
  Titas Chakraborty\thanks{Work done while interning at Motional.} \\
  Carnegie Mellon University\\
  \texttt{titas0602@gmail.com} \\
  \And
  Akshay Bhagat \\
  Motional \\
  \texttt{akshay.bhagat@motional.com} \\
  \AND
  Henggang Cui \\
  Motional \\
  \texttt{henggang.cui@motional.com} \\
}

\begin{document}

\maketitle

\begin{abstract}
Robust motion forecasting of the dynamic scene is a critical component of an autonomous vehicle.
It is a challenging problem due to the heterogeneity in the scene and the inherent uncertainties in the problem.
To improve the accuracy of motion forecasting, in this work, we identify a new consistency constraint in this task, that is \emph{an agent's future trajectory should be coherent with its history observations and visa versa}. To leverage this property, we propose a novel cycle consistency training scheme and define a novel cycle loss to encourage this consistency. In particular, we reverse the predicted future trajectory backward in time and feed it back into the prediction model to predict the history and compute the loss as an additional cycle loss term. Through our experiments on the Argoverse dataset, we demonstrate that cycle loss can improve the performance of competitive motion forecasting models.
\end{abstract}

\section{Introduction}

Motion forecasting in autonomous vehicles is a challenging research problem to solve for the autonomous driving industry. Motion forecasting involves predicting the future trajectories of the agents in a scene given their history trajectories and the scene context, usually in the form of a lane graph. The problem is multi-modal since, given a history, there can be multiple possible futures. For example, at an intersection, an agent can take different possible maneuvers (e.g., straight, right turn, left turn) and follow different speed profiles.

Motion forecasting approaches are highly diverse in their input representations. Several methods, such as \cite{home, multipath}, found success representing the input scene as an image and using convolutional neural networks to generate a scene representation. More recently, vectorized representations used in \cite{hivt, autobots, scenetransformer, wayformer} have found success due to computationally efficient sparse representation and ability to model long-range dependencies. \cite{lanegcn, home, gohome, thomas, dsp} represent lanes as nodes in a graph to make use of the connectivity information. There are several output representations and loss functions in use as well. \cite{dcms, lanegcn} predict an unconstrained regressed output and use a simple regression loss. \cite{home, gohome, thomas} predict an unconstrained heatmap with a cross entropy loss, sample the trajectory endpoint from the heatmap, and they complete the whole trajectory conditioned on the endpoint.

There are a number of consistencies and symmetries inherent to the motion forecasting problem. Recently, there has been an effort to integrate this information, either as an explicit loss or in the model structure to improve the model performance. Certain symmetries such as rotation, translation and scale symmetries have been incorporated into the loss function using data augmentation in various methods \cite{lanegcn, dcms, dsp}. \cite{hivt} incorporates rotation and translation invariance explicitly into the model structure and input representation. \cite{dcms} enforces temporal and spatial consistency constraints into the loss function, which is shown to improve prediction accuracy.

\begin{figure}
    \centering
    \includegraphics[width=1.0\columnwidth]{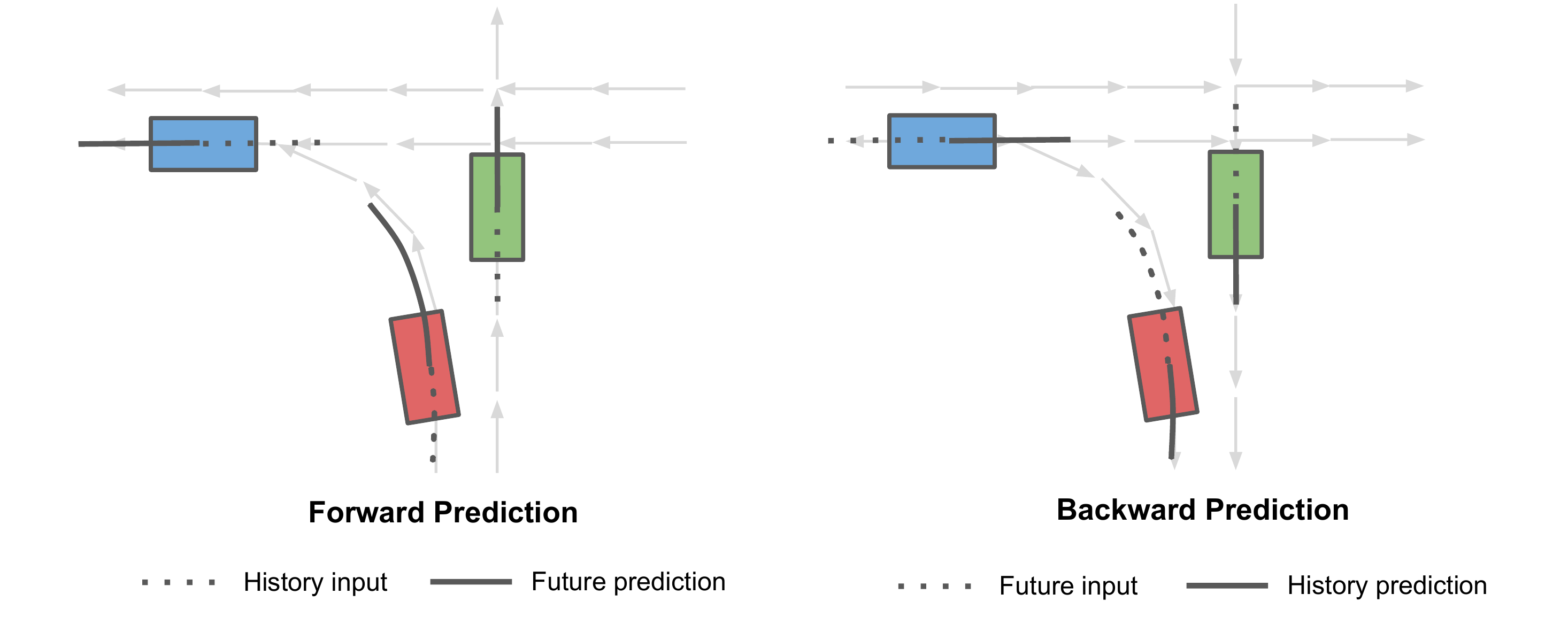}
    \caption{Cycle prediction. In the backward prediction pass, we reverse the trajectories as well as the direction of the road graph, and let the model predict backward in time.}
    \label{fig:cycle-prediction}
\end{figure}

In this work, we propose a novel consistency-based loss called \textit{cycle loss}, which is illustrated in Figure~\ref{fig:cycle-prediction}.
The key motivation of our method is that an agent's future trajectory should be coherent with its past observations, and this coherency should still hold even if we reverse the sequence backward in time.
Motivated by this observation, we propose to add an auxiliary task and loss term in the existing training scheme of the motion forecasting model.
After the model predicts the future trajectory using the history input,
we reverse the predicted future trajectory backward in time and reverse the direction of the road graph, and we feed them back into the prediction model to let it predict the history.
We compute the loss of this auxiliary task as an additional cycle loss term.
Similar to the temporal consistency loss proposed in~\cite{dcms}, our cycle loss method is generic and can be applied to any motion forecasting model.

We summarize our contributions as follows:

\begin{enumerate}
    \item We propose \textit{cycle loss}, a novel training scheme and consistency loss for motion forecasting. This loss explicitly ensures that the future trajectories predicted by the model are coherent with the history observations.
    We also introduce the concept of \textit{ground truth trajectory mixing}, which is necessary for cycle loss to work.
    \item We conduct extensive experimentation on the Argoverse dataset \cite{Argoverse}. We justify the design choices needed for cycle loss to work through ablations and demonstrate that cycle loss can improve the performance of competitive motion forecasting models.
\end{enumerate}

\section{Related Work}

\textbf{Attention-based motion forecasting models:} Attention is used widely in motion forecasting models to fuse diverse features, especially agent and lane features. \cite{lanegcn} proposed a novel form of graph convolution to aggregate short and long-range lane features and used self and cross attention to fuse agent and lane features. \cite{home, gohome, thomas} also use self and cross-attention between agent and lane features. More recently, several transformer-based papers such as \cite{wayformer, autobots, scenetransformer} incorporate temporal attention as well. Some works, such as \cite{hivt}, introduce a novel formulation of attention to enforce spatial and rotational invariance. Other works, such as \cite{dsp}, model the scene using two graph layers and use attention to fuse embeddings of the two layers.
In our experiments, we applied our proposed cycle loss method to two attention-based motion forecasting models, GOHOME~\cite{gohome} and Autobots~\cite{autobots}. We selected them because of their popularity and competitive performance.

\textbf{Consistency-Based Losses:} There are several inherent constraints in the motion forecasting problem that have been discussed in the literature. \cite{dcms} introduced temporal and spatial consistency losses. Temporal consistency enforces that inputs shifted by a small time interval produce similar output trajectories. Spatial consistency enforces that inputs perturbed by a small amount of noise produce similar output trajectories. The loss introduced by \cite{tenet} is the closest to our work in motion prediction of autonomous vehicles. They pass learning embeddings for the future timesteps through a temporal flow network to reconstruct the input. However, since they use a separate model for reconstruction, it is not truly a consistency loss; but rather a method to enrich the learned embeddings. In our work, we use the same model to predict forward and backward in time, making our method a consistency loss on the model. Another difference between our cycle loss method and \cite{tenet} is that we use the predicted trajectories to do backward prediction, while \cite{tenet} uses feature embeddings. \cite{Sun_2020_CVPR} uses a similar approach to ours for human trajectory prediction. They pretrain two models for forward and backward prediction and refine the predictions for both models by jointly training both models with cycle loss. This makes their training procedure lengthy as they have to train three times. Our approach differs since we use the same model for forward and backward prediction making our training end-to-end and therefore much faster than \cite{Sun_2020_CVPR}. This also makes cycle loss a true consistency loss on the model which is not the case for \cite{Sun_2020_CVPR}. The element of \textit{ground truth trajectory mixing} is crucial for our method to work.

\section{Methods}

\subsection{Problem Statement} \label{subsection:probstat}

The general motion forecasting problem involves predicting the future trajectories of agents given the history trajectories and map context (usually represented as a lane graph). Consider that the number of agents in the scene is $A$, of which $P$ are target agents whose trajectories need to be predicted by the model, and $B$ are background agents which are used to provide scene context for the predictions of the target agents. Let the history length be $H$ and the future length to be predicted be $F$. We are given the history trajectory of the target agent $i$ as $\textbf{x}^i= \{x_1^i, \dots , x_H^i \}$ and the history trajectory of background agent $i$ as $\textbf{b}^i= \{b_1^i, \dots , b_H^i \}$ where each element is a 2D array with $x$ and $y$ coordinates.
The map context $\mathcal{M}$ is represented as a directional lane graph, where each node in the graph is a lane segment with a start point and an end point.
We denote the ground-truth future positions of target agent $i$ as $\textbf{y}^i= \{y_1^i, \dots , y_F^i \}$ and background agent $i$ as $\textbf{b}^i_f= \{y_{f1}^i, \dots , y_{fF}^i \}$.
Our task is to predict $K$ future trajectories for each target agent, where the $k$-th future trajectory is denoted as $\textbf{y}^{pik}= \{y_1^{pik}, \dots , y_F^{pik} \}$.

\subsection{Cycle Consistency Training} \label{subsection:cycledesc}

\begin{figure}
    \centering
    \includegraphics[width=1.0\columnwidth]{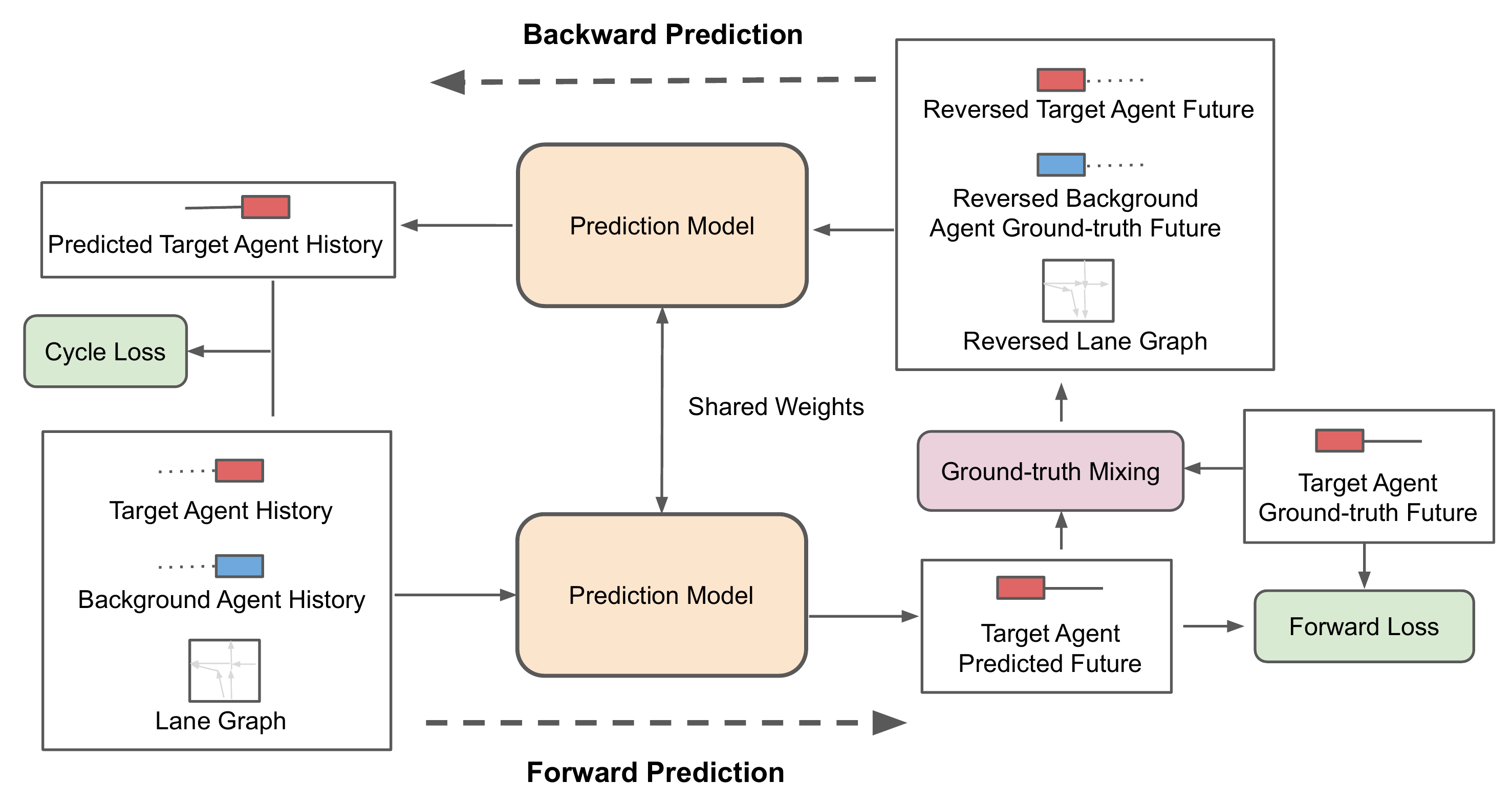}
    \caption{Cycle consistency training architecture}
    \label{fig:cycleloss}
\end{figure}


The overall training scheme with cycle loss is illustrated in Figure~\ref{fig:cycleloss}. We conduct two passes of the model, a normal \textit{forward prediction pass} that is the same as the regular motion forecasting training and a \textit{backward prediction pass} that reverses the trajectories as well as the lane graph backward in time.

In the \textit{forward prediction pass}, we use the input agent histories, $\textbf{x}$ and $\textbf{b}$, along with the map context $\mathcal{M}$, to predict the target agent future trajectories $\textbf{y}^{pk}$.
We use the standard winner-takes-all multimodal prediction loss~\cite{mtp} as the forward loss, which has a mode classification term and a trajectory regression term.
The trajectory regression term is computed on the prediction trajectory that is closest to the ground-truth measured by the final displacement error (FDE).

In the \textit{backward prediction pass}, we reverse the future trajectories of the agents as well as the directions of the road graph, and feed them back to the model to predict the history trajectory.
When the model predicts $K>1$ future trajectories for the target agent, we pick the trajectory that is closest to the ground-truth as measured by the final displacement error (FDE), denoted as $\textbf{y}^{p}$. And we denote the reversed future trajectory as $\textbf{y}^{rp}$ where $\textbf{y}^{rpi} = \{y_F^{pi}, \dots , y_1^{pi} \}$, which is the backward pass input for the target agent.
For the background agents that the model doesn't predict for, we use their ground-truth future trajectories to reverse and get $\textbf{b}^{r}_f$ where $\textbf{b}^{ri}_f = \{y_{fF}^{i}, \dots , y_{f1}^{i} \}$.
We also reverse the map context to obtain $\mathcal{M}^r$.
For the map context, we reverse the connection directions of the lane graph, and we reverse the directions of all the lane segments in the graph.

In the simpler case when $F=H$, we use $\textbf{x}_{rev}=\textbf{y}^{rp}$ and $\textbf{b}_{rev}=\textbf{b}_f^r$, along with the reversed map context $\mathcal{M}^r$, to predict backward in time the target agent history trajectories $\textbf{y}^{pk}_{rev}$.
The predicted history trajectory and the original agent history $\textbf{x}$ are then used to compute our \textit{cycle loss}.
Similar to the forward pass, we use the winner-takes-all approach to compute the cycle loss.
We pick the history trajectory that is closest to the original history as measured by the final displacement error (FDE) to calculate the loss.

$$Cycle \; Loss \; = \frac{1}{HP}\sum_{i=1}^{i=P} \min_{k \in \{1, \dots K\}}\sum_{j=1}^{j=H}||y^{pik}_{rev, H-j} - x_j^i||_2$$

\subsection{Trajectory Truncation or Extension for $F \ne H$}

In many datasets (such as Argoverse), the length of the future prediction trajectory $F$ is longer than the history length $H$, that is, $F > H$.
For those datasets, we need to truncate the future trajectory to length $H$ when passing as the input to the backward pass.
We provide ablations for different choices of truncation parameters in Section~\ref{section:predgtimp}. 

Similarly, in datasets where $F < H$, we need to extend the future trajectory with a part of the agent history to make it length $H$.

\subsection{Ground Truth Trajectory Mixing}
\label{subsection:gt-mixing}

If we use purely the predicted target agent trajectories as the input to the backward pass, there is a possibility that this encourages the model to predict future trajectories that are easier to regress backward but are less accurate. For example, the model might predict purely straight line modes instead of accounting for right or left turns since straight line modes have only velocity uncertainty as opposed to other modes that have maneuver uncertainty as well.

One straightforward way to mitigate this issue is to reduce the weight of the cycle loss term,
but this will limit the improvement offered by cycle loss.
In this work, we find a better approach to addressing this issue is to mix the predicted future trajectories with the ground-truth future trajectories with a probability $p$ to generate a \textit{mixed trajectory}, and use the mixed trajectory as input to the backward pass. This ensures that if the predicted future trajectory is too far from the ground truth, the mixed trajectory will be completely infeasible and will result in poor backward predictions and a high cycle loss.

Mathematically, as we described in Section~\ref{subsection:cycledesc}, after reversing the predicted target agent future trajectories, we obtain $\textbf{y}^{rp} \in \mathcal{R}^{P \times F \times 2}$. We also flip the ground truth target agent future trajectories to obtain $\textbf{y}^r$ where $\textbf{y}^{ri}= \{y_F^{i}, \dots , y_1^{i} \}$. Here, $\textbf{y}^{r} \in \mathcal{R}^{P \times F \times 2}$. We generate a binary mask which we denote as $GTM \in \mathcal{R}^{P \times F \times 2}$ where each element of $GTM$ is generated independently and is $1$ with probability $p$ and $0$ with probability $1-p$. Thus, with ground truth trajectory mixing,  

$$\textbf{x}_{rev} = GTM \otimes \textbf{y}^{rp} + (1 - GTM) \otimes \textbf{y}^{r}$$

\subsection{Implementation Details} \label{subsection:modelarch}

\subsubsection{GOHOME Implementation}

\begin{figure}
    \centering
    \includegraphics[width=1.0\columnwidth]{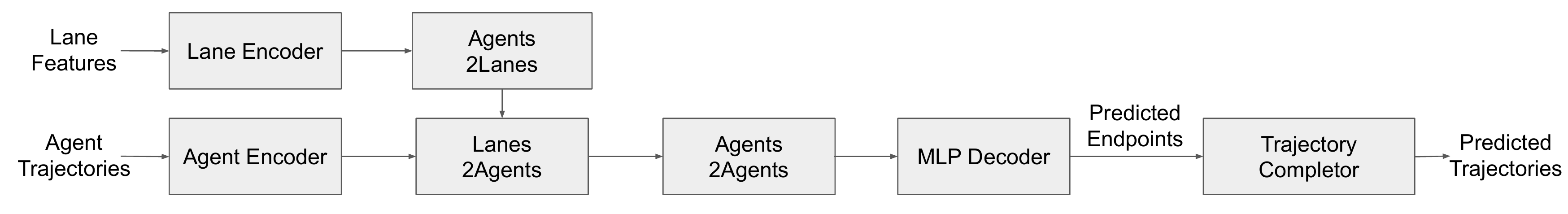}
    \caption{Block diagram for our GOHOME-Mod model}
    \label{fig:gohomeblock}
    \vspace{5px}
\end{figure}

We implemented our cycle consistency training scheme on GOHOME \cite{gohome}, which is a goal-based motion forecasting model using attentions. It first predicts the endpoint of the trajectory and then completes the whole trajectory conditioned on the goal.
Since the original GOHOME model code is not open-sourced, we implemented our own variant of the GOHOME model, and we refer to it as \textit{GOHOME-Mod} for the rest of the paper.

The architecture of our GOHOME-Mod model is illustrated in Figure~\ref{fig:gohomeblock}.
We use 1D CNN + GRU as our encoder in both agent and lane encoders to encode agent trajectories and lane centerlines. We use cross-attention block \textit{Agents2Lanes} to generate agent-aware lane features and use cross-attention block \textit{Lanes2Agents} to generate lane-aware agent features. We then apply self-attention block \textit{Agents2Agents} to produce inter-aware agent features and use a $3$-layer MLP decoder to predict the trajectory endpoints of the target agents. Finally, we use a trajectory regressor module to complete the trajectory conditioned on the predicted endpoint and the agent feature embeddings.
Similar to the other goal-based prediction approaches such as \cite{home, gohome, thomas, dcms}, we train the trajectory regressor conditioned on the ground truth trajectory endpoints. For all attention blocks, we use multi-head attention with $8$ attention heads.

We add our proposed cycle loss as an additional loss term in addition to its original forward losses of GOHOME-Mod.
We refer to the model with cycle consistency loss as \textit{GOHOME-Mod + CL}.


\subsubsection{Autobots Implementation}

We also applied our cycle loss method to the official open-source implementation of Autobots \cite{autobots}, which is a transformer-based motion prediction model. We used the single agent prediction version of the model (Autobots-Ego) since it has better performance on the Argoverse dataset. Same as the GOHOME-Mod + CL model, we added the cycle loss as an additional loss term in addition to its original forward losses.
We refer to the model with cycle consistency loss as \textit{Autobots + CL}.

\section{Evaluation}

\subsection{Datasets}

We use the Argoverse 1.1 dataset \cite{Argoverse}, which is a widely used dataset for motion forecasting to evaluate our method. The Argoverse 1.1 dataset consists of real-world driving scenarios with 205,942 training samples, 39,472 validation samples, and 78,143 test samples. For training and validation samples, the dataset provides a $2$ second history and a $3$ second future trajectories sampled at $10$ Hz. For the test samples, only the $2$ second history is provided. The dataset is collected from two cities, Miami and Pittsburgh. In addition to the agent trajectories, the map information is also provided in the form of a lane graph. It represents lanes as lane centerlines, which is a sequence of points, with lane attributes including turn direction, presence of traffic control, and presence of intersections. To construct the reversed lane features required by cycle consistency, we reverse the turn directions and reverse the order of points in each centerline. Argoverse is a single-agent dataset that only requires the models to predict the trajectories for one single target agent in each frame. Therefore, using our terminology described in Section~\ref{subsection:probstat}, for the Argoverse dataset, $H=20$, $F=30$ and $P=1$.

We use the minADE, minFDE, MR, and DAC metrics, which are standard metrics used by the Argoverse leaderboard~\cite{Argoverse}. minFDEK is the minimum displacement between the last waypoint of the ground truth trajectory and the last predicted waypoint of top K predicted trajectories. minADEK is the minimum average displacement between all waypoints of the ground truth trajectory and all predicted waypoints of top K predicted trajectories. MR-K represents the rate at which minFDEK exceeds $2$ meters over the entire dataset. DACK represents the percentage of the top K predicted trajectories that do not leave the drivable area at any point. For minADE, minFDE, and MR, lower numbers are better. For DAC, higher numbers are better.

\subsection{Experimental Details}

For the GOHOME-Mod model, we train the model for $200$ epochs with an initial learning rate of $10^{-3}$ decayed by $0.5$ every $60$ epochs.

For the Autobots model, we train the model for $150$ epochs with an initial learning rate of $7.5 \times 10^{-4}$ decayed by $0.5$ every $20$ epochs.

Unless otherwise specified, for both models, we use the first $20$ predicted future timesteps to reverse and pass them into the model in the backward prediction pass, and we use a probability of $p=0.5$ for ground truth trajectory mixing.
We set the weight of cycle loss to $2$ for GOHOME-Mod and to $1$ for Autobots.

\subsection{Argoverse Test Results}

\begin{table}[H]
\centering
\begin{tabular}{|c|c|c|c|c|}
\hline
{\bf Model} &  {\bf minFDE6} & {\bf minADE6} & {\bf DAC6} & {\bf MR6} \\ \hline
THOMAS \cite{thomas} & 1.4388 &  0.9423 & 0.9781 & {\bf 0.1038} \\ \hline
TNT \cite{tnt} & 1.4457 &  0.9097 & {\bf 0.9889} & 0.1656 \\ \hline
GOHOME \cite{gohome} & 1.4503 &  0.9425 & 0.9811 & 0.1048 \\ \hline
GOHOME-Mod (ours) & 1.4368 &  0.9426 & 0.9787 & 0.1763 \\ \hline
GOHOME-Mod + CL (ours) & {\bf 1.3896} &  {\bf 0.8949} & 0.9833 & 0.1682 \\ \hline
\end{tabular}
\caption{Results on Argoverse Test Set. The baseline numbers are from the leaderboard.}
\label{Tab:testres}
\end{table}

Table~\ref{Tab:testres} shows the results of our models on the Argoverse test set compared to some standard baselines, including THOMAS \cite{thomas}, TNT \cite{tnt}, and the original GOHOME \cite{gohome} model.
The numbers for the baseline models are directly copied from the leaderboard.
We can see that, by adding cycle loss, our GOHOME-Mod + CL model outperforms all metrics of its corresponding baseline GOHOME-Mod and improves minFDE6 by a significant amount.
This result demonstrates the effectiveness of our cycle loss method.
The result also shows our GOHOME-Mod implementation archives similar performance as the reported numbers of the original GOHOME model on the Argoverse leaderboard, indicating that our re-implementation is a competitive model. 

\subsection{Ablation Studies}

We performed the following ablation studies on the Argoverse validation set.

\subsubsection{Importance of Predicted and Ground-Truth Positions}
\label{section:predgtimp}

\begin{table}[H]
\centering
\begin{tabular}{|c|c|c|c|c|}
\hline
{\bf Model} & {\bf Future input} & {\bf History target} & {\bf ValFDE6} & {\bf ValADE6} \\ \hline
GOHOME-Mod & - & - & 1.163 & 0.767 \\ \hline
GOHOME-Mod + CL & 30-50 & 1-30 & 1.146 & 0.7543 \\ \hline
GOHOME-Mod + CL & 20-40 & 1-20 & \textbf{1.104} & \textbf{0.7369} \\ \hline
\end{tabular}
\caption{Ablation study for the predicted and matched positions used for cycle loss. A position of a-b indicates the positions from the a$^{th}$ timestep to the b$^{th}$ timestep are used for calculating cycle loss. Trajectories in Argoverse $1$ have a total of $50$ timesteps (20 history + 30 future).}
\label{tab:ablation-positions}
\end{table}

In Argoverse, the length of the future trajectory is longer than the length of the history ($H=20$ and $F=30$).
As a result, in the backward prediction pass, we need to pick which part of the future trajectory we feed as the backward input.
In this ablation study, we studied different choices of this parameter, as summarized in Table~\ref{tab:ablation-positions}.

The result shows that the model performs the best when we feed in the first $20$ future waypoints as the backward pass input and use them to predict the $20$ history waypoints. The losses for the remaining part of the backward prediction trajectories are masked out.
We believe the reason why this setting yields the best performance is because the uncertainty in the last $20$ positions is larger compared to the uncertainty in the first $20$ positions, making it tough for the model to obey cycle consistency.

\subsubsection{Importance of Ground Truth Trajectory Mixing}

\begin{table}[H]
\centering
\begin{tabular}{|c|c|c|}
\hline
\textbf{Model} & \textbf{Mixing probability ($p$)} & \textbf{ValFDE6} \\ \hline
GOHOME-Mod & - & 1.163  \\ \hline
GOHOME-Mod + CL & 0 (All Ground Truth Positions) & 1.133 \\ \hline
GOHOME-Mod + CL & 1 (All Predicted Positions) & 1.184 \\ \hline
GOHOME-Mod + CL & 0.5 & \textbf{1.104} \\ \hline
\end{tabular}
\caption{Ablation study for the mixing probability of ground truth positions and predicted positions. The first $20$ predicted waypoints are used to feed into the model, and the $20$ input positions are used for calculating cycle loss, as is found optimal in Section~\ref{section:predgtimp} }
\label{tab:gt-mixing}
\end{table}

In Table~\ref{tab:gt-mixing}, we studied different parameters of ground-truth mixing (described in Section~\ref{subsection:gt-mixing}).
As we can see from the results, mixing ground truth future trajectories with the predicted future trajectories gives us the best FDE on the validation set.
An interesting fact to note is that, when we use all ground truth futures as the backward prediction inputs, it can also reduce the FDE by an amount of $3$ cm over the baseline. Since using all ground truth futures essentially corresponds to augmenting the dataset using the time-reversed data, the result suggests that time-reversing can also be used as an augmentation strategy to enhance the training of the motion forecasting models. 

We believe the main reason why \textit{time flipping augmentation} helps in motion forecasting is that it reduces the imbalances of complex maneuvers in the dataset. For example, if the dataset has more left turns than right turns, \textit{time flipping augmentation} can balance the number of those maneuvers. Similarly, a car might increase its speed or decrease its speed in the future trajectory. If the number of instances of the car speeding up are higher than the car slowing down, that creates an imbalance which \textit{time flipping augmentation} can again correct.

Meanwhile, when we use all predicted futures, its performance is worse than not applying cycle loss at all. This highlights the importance of ground truth trajectory mixing, which ensures that the predicted trajectory does not stray off the ground truth in a bid to minimize the cycle loss.

\subsection{Additional Results on Autobots}

In addition to GOHOME-Mod, we also implemented cycle loss on Autobots~\cite{autobots}. The design choices made were the same as in the GOHOME-Mod model, except that the weight of cycle loss was fixed at $1$ based on our parameter search.

\begin{table}[H]
\centering
\begin{tabular}{|c|c|}
\hline
{\bf Model} &  {\bf minFDE6} \\ \hline
Autobots \cite{autobots}  & 1.114\\ \hline
Autobots + CL & \textbf{1.084} \\ \hline
\end{tabular}
\caption{Autobots Results on Argoverse Validation Set}
\label{tab:autobots-results}
\end{table}

As the results in Table~\ref{tab:autobots-results} show, the addition of cycle loss improves the prediction performance of the open-sourced Autobots model as well.
This result demonstrates that our cycle loss method is generic.

\section{Limitations}

There are several limitations to our proposed consistency loss method.
First, similar to the temporal consistency loss proposed in \cite{dcms}, the cycle loss method incurs overhead at the training time, as it requires two passes of the model during training, nearly doubling the training time.
Second, some of the cycle loss parameters depend on the model architecture and dataset. For example, we find the optimal cycle loss weight is different in GOHOME-Mod and in Autobots.
Third, cycle loss requires the model to be end-to-end differentiable. Some models \cite{gohome, thomas, home, densetnt} rely on sampling a probability distribution as an intermediate step, and more sophisticated designs will be needed in order to apply cycle loss to those models.

\section{Conclusion}

In this work, we propose cycle loss, a novel consistency-based training scheme and loss for motion forecasting. We demonstrated its effectiveness on two competitive motion forecasting models on the Argoverse dataset. The result shows that cycle loss is able to improve the prediction performance of those models by a significant margin.

\clearpage

{
\small

\bibliographystyle{plainnat}
\bibliography{ref.bib}

\begin{thebibliography}{17}
\providecommand{\natexlab}[1]{#1}
\providecommand{\url}[1]{\texttt{#1}}
\expandafter\ifx\csname urlstyle\endcsname\relax
  \providecommand{\doi}[1]{doi: #1}\else
  \providecommand{\doi}{doi: \begingroup \urlstyle{rm}\Url}\fi

\bibitem[Chai et~al.(2020)Chai, Sapp, Bansal, and Anguelov]{multipath}
Yuning Chai, Benjamin Sapp, Mayank Bansal, and Dragomir Anguelov.
\newblock Multipath: Multiple probabilistic anchor trajectory hypotheses for
  behavior prediction.
\newblock In Leslie~Pack Kaelbling, Danica Kragic, and Komei Sugiura, editors,
  \emph{Proceedings of the Conference on Robot Learning}, volume 100 of
  \emph{Proceedings of Machine Learning Research}, pages 86--99. PMLR, 30
  Oct--01 Nov 2020.

\bibitem[Chang et~al.(2019)Chang, Lambert, Sangkloy, Singh, Bak, Hartnett,
  Wang, Carr, Lucey, Ramanan, and Hays]{Argoverse}
Ming-Fang Chang, John~W Lambert, Patsorn Sangkloy, Jagjeet Singh, Slawomir Bak,
  Andrew Hartnett, De~Wang, Peter Carr, Simon Lucey, Deva Ramanan, and James
  Hays.
\newblock Argoverse: 3d tracking and forecasting with rich maps.
\newblock In \emph{Conference on Computer Vision and Pattern Recognition
  (CVPR)}, 2019.

\bibitem[Cui et~al.(2019)Cui, Radosavljevic, Chou, Lin, Nguyen, Huang,
  Schneider, and Djuric]{mtp}
Henggang Cui, Vladan Radosavljevic, Fang-Chieh Chou, Tsung-Han Lin, Thi Nguyen,
  Tzu-Kuo Huang, Jeff Schneider, and Nemanja Djuric.
\newblock Multimodal trajectory predictions for autonomous driving using deep
  convolutional networks.
\newblock In \emph{2019 International Conference on Robotics and Automation
  (ICRA)}, pages 2090--2096. IEEE, 2019.

\bibitem[Gilles et~al.(2021)Gilles, Sabatini, Tsishkou, Stanciulescu, and
  Moutarde]{home}
Thomas Gilles, Stefano Sabatini, Dzmitry Tsishkou, Bogdan Stanciulescu, and
  Fabien Moutarde.
\newblock Home: Heatmap output for future motion estimation.
\newblock In \emph{2021 IEEE International Intelligent Transportation Systems
  Conference (ITSC)}, pages 500--507, 2021.
\newblock \doi{10.1109/ITSC48978.2021.9564944}.

\bibitem[Gilles et~al.(2022{\natexlab{a}})Gilles, Sabatini, Tsishkou,
  Stanciulescu, and Moutarde]{gohome}
Thomas Gilles, Stefano Sabatini, Dzmitry Tsishkou, Bogdan Stanciulescu, and
  Fabien Moutarde.
\newblock Gohome: Graph-oriented heatmap output for future motion estimation.
\newblock In \emph{2022 International Conference on Robotics and Automation
  (ICRA)}, pages 9107--9114, 2022{\natexlab{a}}.
\newblock \doi{10.1109/ICRA46639.2022.9812253}.

\bibitem[Gilles et~al.(2022{\natexlab{b}})Gilles, Sabatini, Tsishkou,
  Stanciulescu, and Moutarde]{thomas}
Thomas Gilles, Stefano Sabatini, Dzmitry Tsishkou, Bogdan Stanciulescu, and
  Fabien Moutarde.
\newblock {THOMAS}: Trajectory heatmap output with learned multi-agent
  sampling.
\newblock In \emph{International Conference on Learning Representations},
  2022{\natexlab{b}}.
\newblock URL \url{https://openreview.net/forum?id=QDdJhACYrlX}.

\bibitem[Girgis et~al.(2022)Girgis, Golemo, Codevilla, Weiss, D'Souza, Kahou,
  Heide, and Pal]{autobots}
Roger Girgis, Florian Golemo, Felipe Codevilla, Martin Weiss, Jim~Aldon
  D'Souza, Samira~Ebrahimi Kahou, Felix Heide, and Christopher Pal.
\newblock Latent variable sequential set transformers for joint multi-agent
  motion prediction.
\newblock In \emph{International Conference on Learning Representations}, 2022.
\newblock URL \url{https://openreview.net/forum?id=Dup_dDqkZC5}.

\bibitem[Gu et~al.(2021)Gu, Sun, and Zhao]{densetnt}
Junru Gu, Chen Sun, and Hang Zhao.
\newblock Densetnt: End-to-end trajectory prediction from dense goal sets.
\newblock In \emph{Proceedings of the IEEE/CVF International Conference on
  Computer Vision}, pages 15303--15312, 2021.

\bibitem[Liang et~al.(2020)Liang, Yang, Hu, Chen, Liao, Feng, and
  Urtasun]{lanegcn}
Ming Liang, Bin Yang, Rui Hu, Yun Chen, Renjie Liao, Song Feng, and Raquel
  Urtasun.
\newblock Learning lane graph representations for motion forecasting.
\newblock In \emph{ECCV}, 2020.

\bibitem[Nayakanti et~al.(2022)Nayakanti, Al-Rfou, Zhou, Goel, Refaat, and
  Sapp]{wayformer}
Nigamaa Nayakanti, Rami Al-Rfou, Aurick Zhou, Kratarth Goel, Khaled~S. Refaat,
  and Benjamin Sapp.
\newblock Wayformer: Motion forecasting via simple \& efficient attention
  networks.
\newblock 2022.
\newblock \doi{10.48550/ARXIV.2207.05844}.
\newblock URL \url{https://arxiv.org/abs/2207.05844}.

\bibitem[Ngiam et~al.(2022)Ngiam, Vasudevan, Caine, Zhang, Chiang, Ling,
  Roelofs, Bewley, Liu, Venugopal, Weiss, Sapp, Chen, and
  Shlens]{scenetransformer}
Jiquan Ngiam, Vijay Vasudevan, Benjamin Caine, Zhengdong Zhang, Hao-Tien~Lewis
  Chiang, Jeffrey Ling, Rebecca Roelofs, Alex Bewley, Chenxi Liu, Ashish
  Venugopal, David~J Weiss, Ben Sapp, Zhifeng Chen, and Jonathon Shlens.
\newblock Scene transformer: A unified architecture for predicting future
  trajectories of multiple agents.
\newblock In \emph{International Conference on Learning Representations}, 2022.
\newblock URL \url{https://openreview.net/forum?id=Wm3EA5OlHsG}.

\bibitem[Sun et~al.(2020)Sun, Zhao, and He]{Sun_2020_CVPR}
Hao Sun, Zhiqun Zhao, and Zhihai He.
\newblock Reciprocal learning networks for human trajectory prediction.
\newblock In \emph{Proceedings of the IEEE/CVF Conference on Computer Vision
  and Pattern Recognition (CVPR)}, June 2020.

\bibitem[Wang et~al.(2022)Wang, Zhou, Zhang, Feng, Lin, Gao, Tang, Zhao, Zhang,
  Guo, Wang, Xu, and Zhang]{tenet}
Yuting Wang, Hangning Zhou, Zhigang Zhang, Chen Feng, Huadong Lin, Chaofei Gao,
  Yizhi Tang, Zhenting Zhao, Shiyu Zhang, Jie Guo, Xuefeng Wang, Ziyao Xu, and
  Chi Zhang.
\newblock Tenet: Transformer encoding network for effective temporal flow on
  motion prediction.
\newblock 2022.
\newblock URL \url{https://arxiv.org/abs/2207.00170}.

\bibitem[Ye et~al.(2022)Ye, Xu, Xu, Cao, and Chen]{dcms}
Maosheng Ye, Jiamiao Xu, Xunnong Xu, Tongyi Cao, and Qifeng Chen.
\newblock Dcms: Motion forecasting with dual consistency and
  multi-pseudo-target supervision.
\newblock 2022.

\bibitem[Zhang et~al.(2021)Zhang, Li, Chen, and Shen]{dsp}
Lu~Zhang, Peiliang Li, Jing Chen, and Shaojie Shen.
\newblock Trajectory prediction with graph-based dual-scale context fusion.
\newblock \emph{arXiv preprint arXiv:2111.01592}, 2021.

\bibitem[Zhao et~al.(2020)Zhao, Gao, Lan, Sun, Sapp, Varadarajan, Shen, Shen,
  Chai, Schmid, Li, and Anguelov]{tnt}
Hang Zhao, Jiyang Gao, Tian Lan, Chen Sun, Benjamin Sapp, Balakrishnan
  Varadarajan, Yue Shen, Yi~Shen, Yuning Chai, Cordelia Schmid, Congcong Li,
  and Dragomir Anguelov.
\newblock Tnt: Target-driven trajectory prediction.
\newblock In \emph{CoRL}, 2020.

\bibitem[Zhou et~al.(2022)Zhou, Ye, Wang, Wu, and Lu]{hivt}
Zikang Zhou, Luyao Ye, Jianping Wang, Kui Wu, and Kejie Lu.
\newblock Hivt: Hierarchical vector transformer for multi-agent motion
  prediction.
\newblock In \emph{Proceedings of the IEEE/CVF Conference on Computer Vision
  and Pattern Recognition (CVPR)}, pages 8823--8833, June 2022.

\end{thebibliography}
}

\end{document}